%% file: main.tex
\documentclass[10pt,twocolumn,letterpaper]{article}

\usepackage{iccv}              

\input{preamble}

\definecolor{iccvblue}{rgb}{0.21,0.49,0.74}
\usepackage[pagebackref,breaklinks,colorlinks,allcolors=iccvblue]{hyperref}

\def\paperID{9816} 
\def\confName{ICCV}
\def\confYear{2025}

\title{Data-Efficient Generalization for Zero-shot Composed Image Retrieval}

\author{Zining Chen$^{1,3}$, Zhicheng Zhao$^{1}$\footnotemark[1],\enspace Fei Su$^{1}$, Xiaoqin Zhang$^{2}$, Shijian Lu$^{3}$\footnotemark[1]\\
$^{1}$ School of Artificial Intelligence, Beijing University of Posts and Telecommunications, China \\
$^{2}$ College of Computer Science and Technology, Zhejiang University of Technology, China \\
$^{3}$ College of Computing and Data Science, Nanyang Technological University, Singapore \\
}

\begin{document}
\maketitle
\input{sec/0_abstract}    
\input{sec/1_intro}
\input{sec/2_formatting}
{
    \small
    \bibliographystyle{ieeenat_fullname}
    \bibliography{main}
}

\end{document}

%% file: preamble.tex
\usepackage{multirow}
\usepackage{float}
\usepackage{subcaption}
\usepackage{amsthm,amsmath,amssymb}
\usepackage{mathrsfs}

%% file: sec/0_abstract.tex
\begin{abstract}
Zero-shot Composed Image Retrieval (ZS-CIR) aims to retrieve the target image based on a reference image and a text description without requiring in-distribution triplets for training. One prevalent approach follows the vision-language pretraining paradigm that employs a mapping network to transfer the image embedding to a pseudo-word token in the text embedding space. However, this approach tends to impede network generalization due to modality discrepancy and distribution shift between training and inference. To this end, we propose a Data-efficient Generalization (DeG) framework, including two novel designs, namely, Textual Supplement (TS) module and Semantic-Set (S-Set). The TS module exploits compositional textual semantics during training, enhancing the pseudo-word token with more linguistic semantics and thus mitigating the modality discrepancy effectively. The S-Set exploits the zero-shot capability of pretrained Vision-Language Models (VLMs), alleviating the distribution shift and mitigating the overfitting issue from the redundancy of the large-scale image-text data. Extensive experiments over four ZS-CIR benchmarks show that DeG outperforms the state-of-the-art (SOTA) methods with much less training data, and saves substantial training and inference time for practical usage.
\end{abstract}

%% file: sec/1_intro.tex
\vspace{-0.4cm}
\begin{figure}
\centering
  \setlength{\belowcaptionskip}{-3pt}
  \includegraphics[width=0.95\linewidth,height=0.3\textheight]{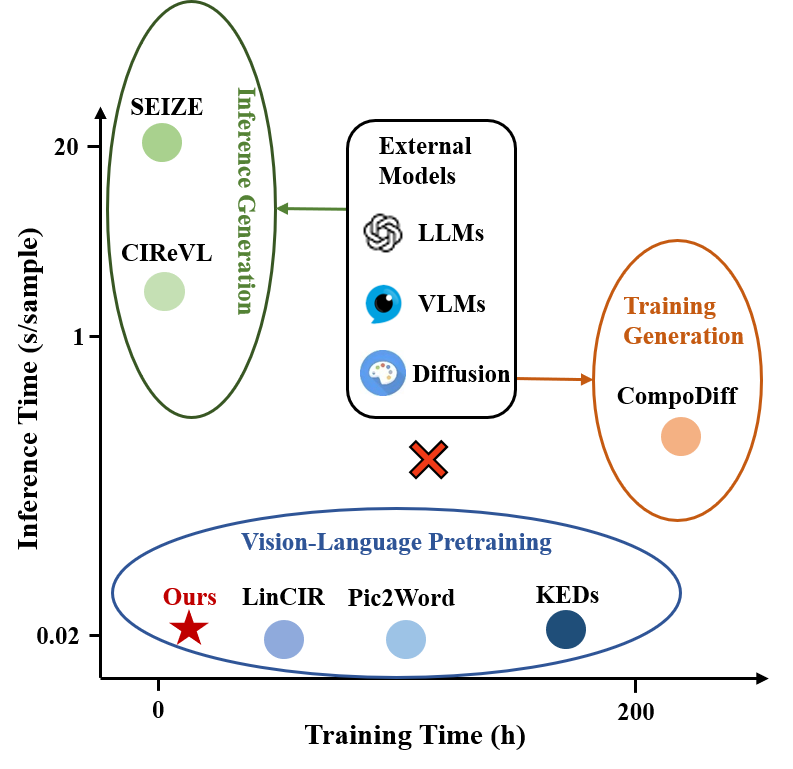}
    \caption{Comparison of training and inference time between different paradigms. Previous methods from vision-language pretraining paradigm requires no external models, but consumes high training costs due to large-scale image-text dataset, while methods from the triplet generation paradigm either consume huge training overhead or inference latency due to training and inference generation by large external models.}
  \label{fig:time}
  \vspace{-\baselineskip}
\end{figure}
\section{Introduction}
\label{sec:intro}
\quad Composed Image Retrieval (CIR) aims to retrieve the target image given a reference image and a textual description for the specific modification of the reference image. Unlike traditional image-based \cite{quan2019auto,chen2023cluster} or text-based retrieval \cite{mishra2013image,alikhani2022cross}, CIR task concentrates on the fusion of visual and textual information, and thus achieves customized retrieval with more flexible and meticulous control from both linguistic and visual perspective. Thanks to its broad application in practical scenarios such as e-commerce, CIR has recently attracted increasing attention from both academic and industrial communities. 

Most previous studies focus on the supervised CIR task that directly uses in-distribution triplets for training, each of which includes a reference image, a textual description and a target image \cite{baldrati2022conditioned,baldrati2022effective,lee2021cosmo,liu2021image,delmas2022artemis,goenka2022fashionvlp,wen2023self,han2023fame}. Leveraging Contrastive Language-Image Pretraining (CLIP) \cite{radford2021learning}, these methods have achieved impressive performance on natural image datasets \cite{liu2021image,baldrati2023zero} and fashion image datasets \cite{wu2021fashion}. However, the collection of the triplets in supervised CIR is laborious and time-consuming. Additionally, the model trained on the triplets often exhibits poor generalization towards novel domains, limiting its scalability and applicability for wide application. 

\begin{figure}
\centering
  \setlength{\belowcaptionskip}{-2pt}
  \includegraphics[width=0.93\linewidth,height=0.3\textheight]{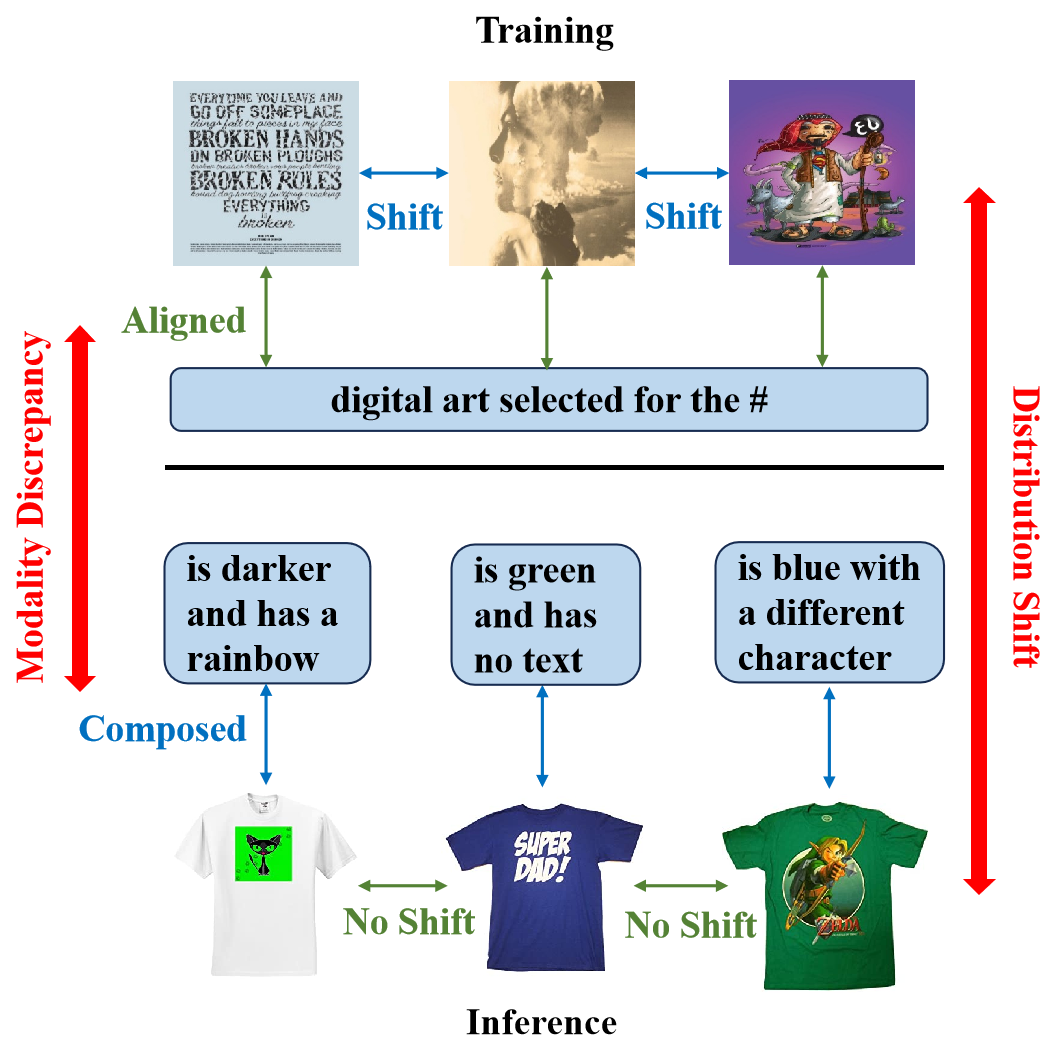}
    \caption{Comparison between the data from the training and inference set. For the modality discrepancy, the visual and textual information during training is aligned, while compositional during inference. Meanwhile, the distribution shift of both image and text between the training and inference dataset is substantial.}
  \label{fig:motiv}
  \vspace{-\baselineskip}
\end{figure}

Zero-shot Composed Image Retrieval (ZS-CIR), which learns without requiring the in-distribution triplets, has been explored to address the issues of supervised CIR. Existing methods can be mainly categorized into two aspects, including triplet generation \cite{liu2023zero,gu2023compodiff,karthik2023vision,zhang2024magiclens,yang2024semantic}, and vision-language pretraining \cite{saito2023pic2word,baldrati2023zero,tang2024context,suo2024knowledge,gu2024language,karthik2023vision}. Triplet generation synthesizes pseudo triplets by leveraging large external models, including diffusion model \cite{rombach2022high}, Large Language Models (LLMs) \cite{brown2020language} and Vision-Language Models (VLMs) \cite{li2022blip}, but
the generation process is computational intensive for both training \cite{liu2023zero,gu2023compodiff} and inference \cite{karthik2023vision,yang2024semantic}, as illustrated in Figure \ref{fig:time}. Vision-Language Pretraining utilizes image-text pairs to pretrain a mapping network, which excels in inference costs \cite{saito2023pic2word,tang2024context,suo2024knowledge,gu2024language}, but still faces several dilemmas from the image-text data, as shown in Figure \ref{fig:motiv}. Specifically, large-scale image-text data is prone to overfitting by nature (\textit{e.g.,} different images may have identical captions) and thus suffer from distribution shift between training and inference data. Besides, the task difference between training and inference, \textit{i.e.,} image and text are aligned during training but compositional during inference, causes modality discrepancy for compositional understanding.

Comprehensively, we propose a Data-efficient Generalization (DeG) framework for effective and efficient ZS-CIR. Specifically, DeG first designs a Textual Supplement (TS) module that alleviates the modality discrepancy by enriching the semantics of the pseudo-word token based on the text descriptions. Secondly, a Semantic Set (S-Set) is introduced to exploit the zero-shot capabilities of pretrained vision-language models to mitigate the overfitting issue. With the two designs, DeG can capture a more semantic and robust pseudo-word token without extra operations from any external models during inference. Meanwhile, DeG employs merely a small portion of the training data as used by SOTA vision-language pretraining methods without any external databases, saving huge training costs. Extensive experiments show that DeG achieves superior retrieval performance on four types of ZS-CIR datasets with much reduced training data and computational costs. Overall, our contributions can be summarized as follows:

\begin{itemize}
    \item We identify two key challenges in ZS-CIR, namely, modality discrepancy and distribution shift, and propose a Data-efficient Generalization (DeG) framework that resolves the challenges to improve the performance of ZS-CIR task.
    \item A novel Textual Supplement (TS) module is designed to provide compositional textual information to improve the pseudo-word token, enhancing compositional understanding and mitigating the modality discrepancy effectively. 
    \item A Semantic Set (S-Set) is introduced to exploit the zero-shot capability of pretrained VLMs with novel training objectives, alleviating the overfitting and thus mitigating the distribution shift effectively.
    \item Experimental results on four ZS-CIR datasets manifest the superior performance of our method with less training data, and the comparison of training and inference time further shows the efficiency and effectiveness. 
\end{itemize}

%% file: sec/2_formatting.tex
\section{Related Work}
\subsection{Composed Image Retrieval} 
\quad Composed Image Retrieval (CIR) intends to retrieve the target image using a relative image and a text description that modifies the content. Recently, with the development of vision-language multi-modal learning, CIR task has attracted wide attention. Many supervised CIR methods \cite{baldrati2022conditioned,baldrati2022effective,lee2021cosmo,liu2021image,delmas2022artemis,goenka2022fashionvlp,wen2023self,han2023fame} have emerged to train on manually-annotated triplets from diverse visual domains, such as natural \cite{liu2021image,baldrati2023zero} and fashion images \cite{wu2021fashion}. These methods mostly focus on the fusion of visual and textual information, and enable the joint embedding closer to the target image embedding. However, the huge expense and time consumption on the triplets hinder the practical usage of these methods, and training on specific dataset deteriorates the generalization on other datasets. 

Therefore, Zero-shot Composed Image Retrieval (ZS-CIR) has recently been proposed to resolve these deficiencies. Currently, two types of work are proposed to tackle ZS-CIR, including triplet generation \cite{liu2023zero,gu2023compodiff,karthik2023vision,zhang2024magiclens,yang2024semantic}, and vision-language pretraining \cite{saito2023pic2word,baldrati2023zero,tang2024context,suo2024knowledge,gu2024language,karthik2023vision}. The former mostly utilizes pretrained generative models \cite{ho2020denoising}, large language models \cite{brown2020language} or vision-language models \cite{li2022blip} to construct synthesized triplets. The latter constructs the pseudo-word token to transfer the image feature into the text embedding space. \cite{saito2023pic2word,baldrati2023zero} are fundamental works to introduce the image mapping network for the pseudo-word token. \cite{tang2024context,suo2024knowledge} improves the mapping network with structural designs. \cite{gu2024language} discovers the limitation of the input texts and proposes to simply utilize language to train the mapping network.

\subsection{Out-of-Distribution Generalization}
\quad Out-of-Distribution (OOD) generalization aims to train a model using In-Distribution (ID) data and generalize to unknown OOD scenarios. Traditional OOD generalization methods can be broadly categorized into three aspects, including data augmentation \cite{xu2021fourier,zhou2021domain}, domain-invariant learning \cite{sun2016deep,shankar2018generalizing, lv2022causality}, and learning strategies \cite{li2018domain, huang2020self,cha2021swad,chen2023instance}. Recently, with the emergence of Contrastive Language-Image Pretraining (CLIP), many approaches concentrate on mining the zero-shot capability of CLIP to resolve the OOD issue \cite{cha2022domain,shu2023clipood,cho2023promptstyler,chen2024practicaldg}. Moreover, multi-modal learning techniques are proposed to tackle the OOD issue \cite{dong2023simmmdg,dong2024towards}, including audio, image and text modalities, etc. Nevertheless, these methods mostly tackle uni-modal tasks, such as image classification \cite{xu2021fourier,lv2022causality} and semantic segmentation \cite{lee2022wildnet,zhao2022style}. Recently, several works on OOD generalization have also addressed the cross-modal tasks, \textit{e.g.,} image captioning \cite{ren2023crossing}, visible-infrared person re-identification \cite{wang2019learning}. However, the generalization issue from multi-modal tasks still remains largely unexplored. As the ZS-CIR task assumes the zero-shot setting, we attempt to exploit the generalization from the training data for improvement.

\subsection{Vision-Language Pretraining} 
\quad Vision-Language Pretraining have achieved great advancements on large-scale image-text datasets \cite{jia2021scaling,lei2021less,radford2021learning}. The CLIP \cite{radford2021learning} establishes the foundation to leverage contrastive learning on large-scale image-text pairs. Later, the emergence of more VLMs, such as ALBEF \cite{li2021align}, Coca \cite{yu2022coca}, BLIP \cite{li2022blip}, have achieved remarkable performance. As VLMs possess strong zero-shot capability, many works exploit these models for various zero-shot tasks, such as zero-shot recognition \cite{zhang2022tip,goyal2023finetune} and open-vocabulary segmentation \cite{zhou2022extract,lan2024clearclip}. \cite{cohen2022my} has first proposed to learn mapping networks to project image and text embeddings into text and image space, and several works have improved it from different objectives on diverse tasks \cite{gal2022image,li2023decap}. However, this paper observes the misalignment between training and inference in the ZS-CIR task, and aims to supplement more semantics to the pseudo-word token with novel objectives.

\section{Method}
\label{sec:method}
\begin{figure*}
\includegraphics[width=0.96\linewidth,height=0.32\textheight]{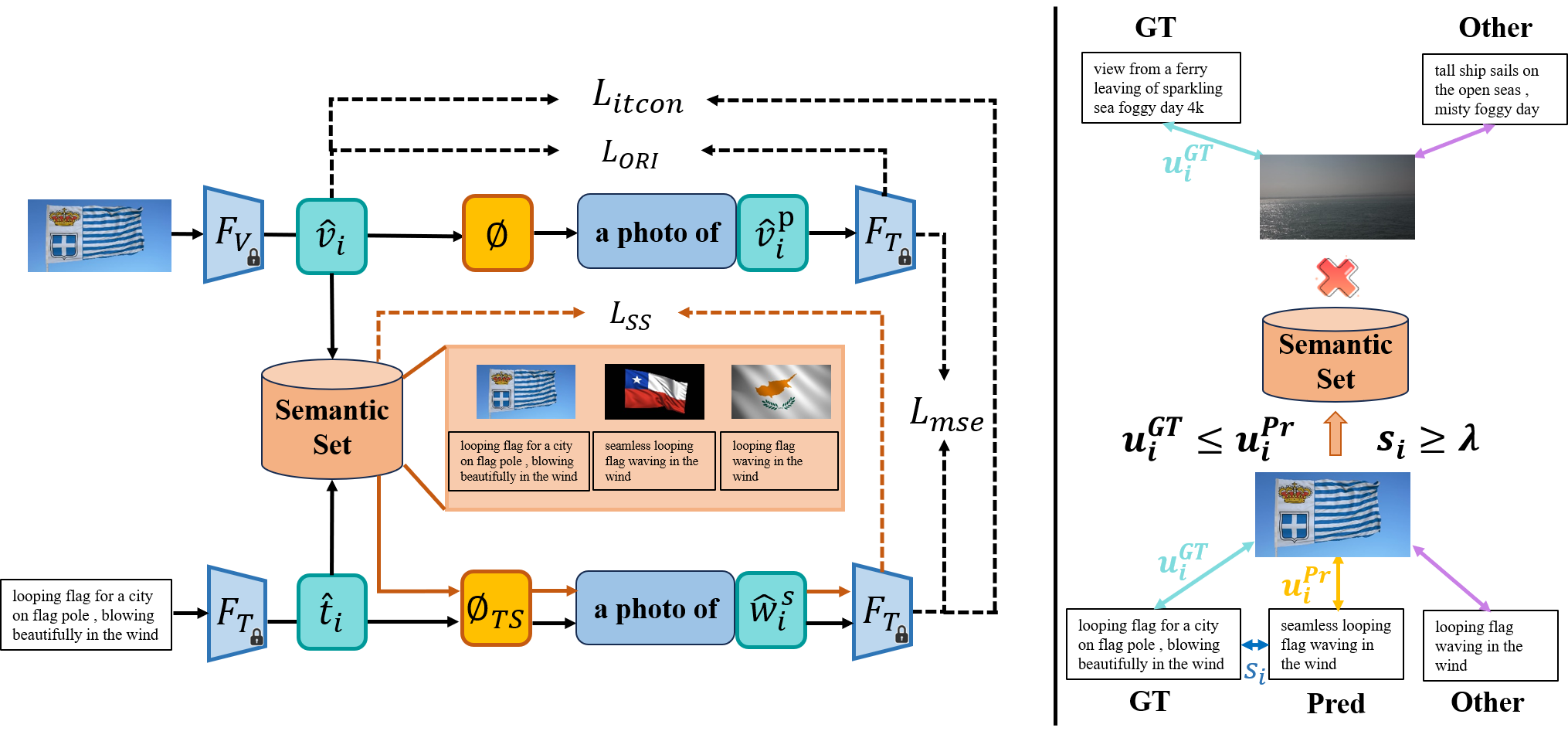}
    \caption{The overall framework of our method DeG. The left part presents the proposed modules, including the Textual Supplement (TS) module and the Semantic Set (S-Set) with novel training objectives. The right part illustrates how the S-Set is selected with two conditions, where the CLIP predicted caption for the image is incorrect but has a similar ground truth caption. }
  \label{fig:Framework}
\vspace{-\baselineskip}
\end{figure*}

\quad In this section, we first introduce the preliminaries of the CLIP model and the ZS-CIR task in Section \ref{sec:3.1}. Then, a detailed description of our method Data-efficient Generalization (DeG) is presented in Figure \ref{fig:Framework}. DeG focuses on enhancing the compositional understanding during training via the Textual Supplement (TS) module in Section \ref{sec:3.2}, and exploiting the generalization ability in the CLIP model by excavating a Semantic Set (S-Set) in \ref{sec:3.3}. Finally, we introduce the inference process of DeG in Section \ref{sec:3.4}.

\subsection{Preliminaries}
\label{sec:3.1}

\quad \textbf{Contrastive Image-Language Pretraining (CLIP) \cite{radford2021learning}.} The CLIP model is trained on an extensive dataset $\mathcal{D}=\left\{(x_i, t_i)\right\}_{i=1}^N$ to achieve alignment between visual and textual modality. For each pair-wise image $x_i$ and tokenized text $t_i$, a parameterized image encoder $F_I$ and text encoder $F_T$ are employed to extract the image embedding $v_i=F_I(x_i)$ and the text embedding $w_i=F_T(t_i)$. Then, the normalized image embedding $\hat{v}_i=\frac{v_i}{||v_i||}$ and text embedding $\hat{w}_i=\frac{w_i}{||w_i||}$ are used to calculate the cosine similarity, and the contrastive loss is utilized as the training objectives to update the image and text encoders,
\begin{equation}
    \mathcal{L}_{con} = \mathcal{L}_{I2T} + \mathcal{L}_{T2I}
\end{equation}
Each item can be defined as,
\begin{equation}
    \mathcal{L}_{I2T} = -\frac{1}{N_B}\sum_{i=1}^{N_B}log\frac{exp(\hat{v}_i^T \cdot \hat{w}_i /\tau)}{\sum_{j=1}^{N_B}exp(\hat{v}_i^T \cdot \hat{w}_j /\tau)}
\end{equation}
\begin{equation}
    \mathcal{L}_{T2I} = -\frac{1}{N_B}\sum_{i=1}^{N_B}log\frac{exp(\hat{w}_i \cdot \hat{v}_i^T /\tau)}{\sum_{j=1}^{N_B}exp(\hat{w}_i \cdot \hat{v}_j^T /\tau)}
\end{equation}
where $\mathcal{B}$ denotes the batch with $N_B$ image-text pairs, $\tau$ is the hyper-parameter on temperature.

\textbf{Zero-shot Composed Image Retrieval (ZS-CIR).} The mainstream of ZS-CIR methods adopt the pseudo-word token technique that uses a mapping module $\phi$ to transfer the image embedding to a pseudo-word token $\hat{v}_i^p=\phi(\hat{v}_i)$. Then, the pseudo-word token $\hat{v}_i^p$ is constructed into the prompt $P_i^p$ as ``a photo of [$\hat{v}_i^p$]", and input into the text encoder to obtain the composed embedding $\hat{V}_i^p=F_T(P_i^p)$. The contrastive loss function is designed to maximize the similarity between the pair-wise composed embedding and the image embedding. 
\begin{equation}
    \mathcal{L}_{I2C} = -\frac{1}{N_B}\sum_{i=1}^{N_B}log\frac{exp(\hat{v}_i^T \cdot \hat{V}_i^p /\tau)}{\sum_{j=1}^{N_B}exp(\hat{v}_i^T \cdot \hat{V}_j^p /\tau)}
\end{equation}
\begin{equation}
    \mathcal{L}_{C2I} = -\frac{1}{N_B}\sum_{i=1}^{N_B}log\frac{exp((\hat{V}_i^p)^T \cdot \hat{v}_i /\tau)}{\sum_{j=1}^{N_B}exp((\hat{V}_i^p)^T \cdot \hat{v}_j /\tau)}
\end{equation}
During inference, each reference image is encoded and fed into the mapping module to obtain the pseudo-word token [\$]. Then, the composed retrieval can be simplified to text-to-image retrieval with a prompt of ``a photo of [\$] that [$cond$]", where [$cond$] is the text description. However, this line of work overlooks the importance of textual information. LinCIR observes the drawback and proposes language-only training, while KEDs introduces an extra database for knowledge enhancement.

\subsection{Textual Supplement}
\label{sec:3.2}
\quad Despite the success of recent ZS-CIR methods, we observe that prior works neglect the modality discrepancy between training and inference. As the inference process focuses on the compositional understanding between the visual and textual modality, the image-only or text-only contrastive losses during training contradict with the ability of compositional understanding. 

To overcome the above issue, we propose a Textual Supplement (TS) module, aiming to learn the complementary information from the captions via novel training objectives. Firstly, we follow previous work \cite{saito2023pic2word} to learn a mapping network $\phi$ with the training objectives $L_{ORI}=\mathcal{L}_{I2C} + \mathcal{L}_{C2I}$. Then, as illustrated in Figure \ref{fig:Framework}, the text encoder encodes the tokenized caption to obtain the text embedding as $\hat{w}_i$, which is input into the TS module for the complementary textual token, $\hat{w}_i^s=\phi_{TS}(\hat{w}_i)$. We design the textual supplement training objectives $\mathcal{L}_{TS}$ to enhance the compositional understanding capability of the network. Concretely, the complementary textual token $\hat{w}_i^s$ is similarly inserted into the prompt $P_i^s$ as ``a photo of [$\hat{w}_i^s$]", and then input into the text encoder to obtain the complementary textual embedding $\hat{W}_i^s=F_T(P_i^s)$. We force $\hat{W}_i^s$ to learn similar semantics as the original image embedding $\hat{v}_i$,
\begin{equation}
    \mathcal{L}_{I2CT} = -\frac{1}{N_B}\sum_{i=1}^{N_B}log\frac{exp(\hat{v}_i^T \cdot \hat{W}_i^s /\tau)}{\sum_{j=1}^{N_B}exp(\hat{v}_i^T \cdot \hat{W}_j^s /\tau)}
\label{equ:i2ct}
\end{equation}
\begin{equation}
    \mathcal{L}_{CT2I} = -\frac{1}{N_B}\sum_{i=1}^{N_B}log\frac{exp((\hat{W}_i^s)^T \cdot \hat{v}_i /\tau)}{\sum_{j=1}^{N_B}exp((\hat{W}_i^s)^T \cdot \hat{v}_j /\tau)}
\label{equ:ct2i}
\end{equation}
\begin{equation}
    \mathcal{L}_{itcon} = \mathcal{L}_{I2CT} + \mathcal{L}_{CT2I}
\end{equation}
Moreover, as the image embedding is projected as a pseudo-word token via the original mapping network $\phi$, the alignment between the pseudo-word embedding $\hat{V}_i^p$ and the complementary textual embedding $\hat{W}_i^s$ should be achieved. Thus, we use the Mean Square Error (MSE) loss function to minimize the gap between the two embeddings from visual and textual modality,
\begin{equation}
    \mathcal{L}_{mse} = ||\hat{V}_i^p - \hat{W}_i^s||_2
\end{equation}
And the overall textual supplement loss function can be defined as, 
\begin{equation}
    \mathcal{L}_{TS} = \mathcal{L}_{itcon} + \alpha \mathcal{L}_{mse}
\end{equation}
where $\alpha$ is a balancing hyper-parameter between two loss functions.

\subsection{Semantic Set}
\label{sec:3.3}
\quad The TS module is capable to refine the pseudo-word token with complementary semantics from text, reducing the modality discrepancy between training and inference on compositional understanding. However, the distribution shift on both image and text between training and inference remains a dilemma. We rethink how to avoid the overfitting issue on large-scale image-text pairs, and excavate the generalization ability of CLIP for robust representations. As CLIP possesses strong zero-shot capability, the invariant property of CLIP can be excavated from the zero-shot prediction of certain samples. Thus, we attempt to establish a Semantic Set (S-Set) that consists of affluent semantic samples from each batch $\mathcal{B}$. 

Concretely, we notice that for each batch consisting of $N_B$ image-text pairs $\left\{(x_i, t_i)\right\}_{i=1}^{N_B}$, the similarity between each image embedding $\hat{v}_i$ and all text embeddings $\left\{\hat{w}_i\right\}_{i=1}^{N_B}$ represents the uncertainty of the image,
\begin{equation}
    u_i = \frac{exp(\hat{v}_i^T \cdot \hat{w}_i /\sigma)}{\sum_{j=1}^{N_B}exp(\hat{v}_i^T \cdot \hat{w}_j /\sigma)}
\end{equation}
where $\sigma$ is set to 0.01 by default. The uncertainty of each image reflects how CLIP measures the possibility between the image embedding and all text embeddings. If even CLIP mistakenly predicts an incorrect caption for the image, the fine-grained semantics of which is affluent that should be further exploited. Therefore, we first select the semantic set based on $u_i$, and the function can be defined as,
\begin{equation}
    M_f(u_i) = \left\{
    \begin{aligned}
    & 1, \quad & MaxInd(u_i) \neq i \\
    & 0, \quad & MaxInd(u_i) = i \\
    \end{aligned}
    \right.
\end{equation}
where $MaxInd(\cdot)$ is a common function that extracts the index with the largest value, aiming to select the one whose similarity at the GT position $u_i^{GT}$ is smaller than the predicted position $u_i^{Pr}$. Then, after obtaining the images that cause the inconsistency, we further select the images with similar captions. A threshold $\lambda$ is defined to filter the images with higher similarity between the predicted text embedding and ground truth text embedding $s_{i}=\langle{\hat{w}_{MaxInd(u_i)}, \hat{w}_{i}}\rangle$,
\begin{equation}
    M_s(u_i) = \left\{
    \begin{aligned}
    & 1, & s_{i} \geq \lambda \\
    & 0, & s_{i} < \lambda \\
    \end{aligned}
    \right.
\end{equation}
As a result, the selected images are all incorrectly predicted by CLIP, but have similar captions to cause ambiguity,
\begin{equation}
M_i = M_f(u_i) \times M_s(u_i)
\end{equation}
where $M_i=1$ means the image $x_i$ is selected into the semantic set, and vice versa. Thus, the semantic set contains image-text pairs with high similarity that causes misunderstanding of CLIP, which is essential for fine-grained semantics excavation. The loss function is designed as follows,
\begin{equation}
    \mathcal{L}_{SI2CT} = -\frac{1}{N_S}\sum_{i=1}^{N_S} log\frac{exp(\hat{v}_i^T \cdot \hat{W}_i^s /\tau)}{\sum_{j=1}^{N_S}exp(\hat{v}_i^T \cdot \hat{W}_j^s /\tau)}
\label{equ:ssi2t}
\end{equation}
\begin{equation}
    \mathcal{L}_{SCT2I} = -\frac{1}{N_S}\sum_{i=1}^{N_S} log\frac{exp((\hat{W}_i^s)^T \cdot \hat{v}_i /\tau)}{\sum_{j=1}^{N_S}exp((\hat{W}_i^s)^T \cdot \hat{v}_j /\tau)}
\label{equ:sst2i}
\end{equation}
where $N_S$ is the number of samples in the semantic set. The overall training objectives on the semantic set is,
\begin{equation}
    \mathcal{L}_{SS} = \mathcal{L}_{SI2CT} + \mathcal{L}_{SCT2I}
\end{equation}
The loss function $\mathcal{L}_{SS}$ is capable to enhance the capability of fine-grained information mining via the guidance of CLIP, which is significant during inference process. Meanwhile, the zero-shot capability of CLIP is exploited via the image selection process to construct the semantic set. The final training objectives of DeG can be defined as,
\begin{equation}
     \mathcal{L}_{DeG} = \mathcal{L}_{ORI} + \mathcal{L}_{TS} + \beta \mathcal{L}_{SS}
\end{equation}
where $\beta$ is a hyper-parameter to balance the ratio of different losses.

\subsection{Inference Process}
\label{sec:3.4}
\begin{figure}
  \includegraphics[width=1\linewidth,height=0.11\textheight]{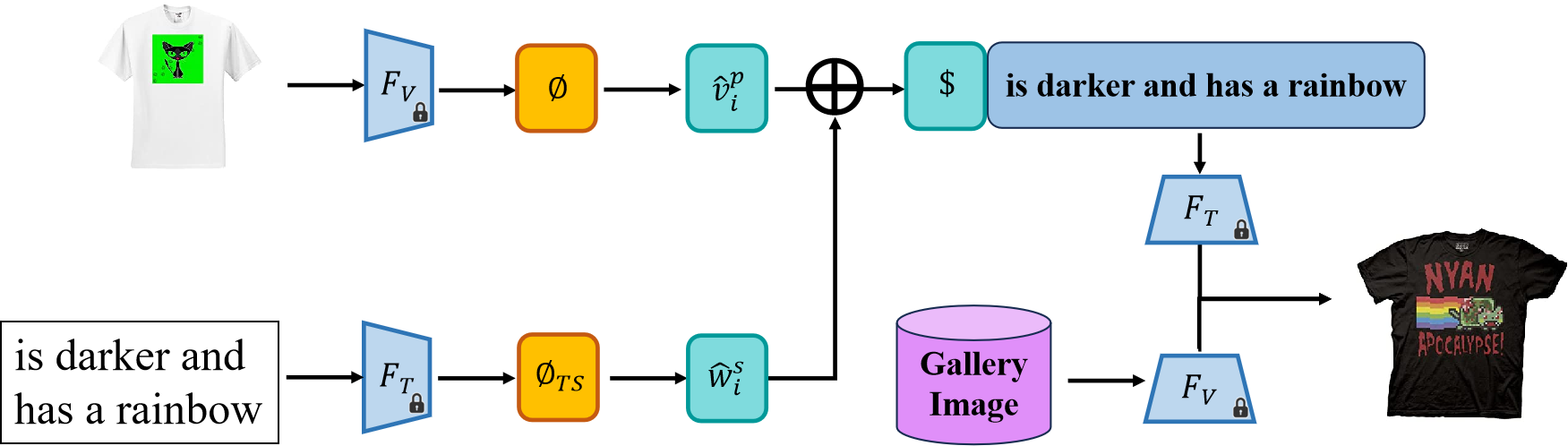}
    \caption{The inference process of our method DeG.}
  \label{fig:infer}
    \vspace{-\baselineskip}
\end{figure}

\quad As shown in Figure \ref{fig:infer}, during inference, DeG first produces the pseudo-word token $\hat{v}_i^p$ from the reference image via the conventional mapping network $\phi$. Then, the text description is constructed to a prompt ``a photo of [$cond$]", and is input into the TS module $\phi_{TS}$ to generate the textual supplement token $\hat{w}_i^s$. The overall semantic token can be mixed for retrieval as,
\begin{equation}
    [\$] = \gamma \times \hat{v}_i^p + (1-\gamma) \times \hat{w}_i^s 
\end{equation}
where $\gamma$ is the mixing ratio from original mapping network and the TS module. Then, ``a photo of [\$] that [$cond$]" is input into the text encoder to retrieve the most similar target image as the result. 

\begin{table*}[ht]
\caption{Results on Fashion-IQ validation set. * indicates the supervised method Combiner with ResNet50x4 as the backbone.}
\vspace{-\baselineskip}
\label{tab:fashion}
\begin{center}
\setlength{\tabcolsep}{6pt}
\begin{tabular}{c|c|cc|cc|cc|cc|cc}
\hline
\multirow{2}{*}{Supervision} & \multirow{2}{*}{Methods} & \multicolumn{2}{c|}{Amount} & \multicolumn{2}{c|}{Dress} & \multicolumn{2}{c|}{Shirt} & \multicolumn{2}{c|}{TopTee} & \multicolumn{2}{c}{Average}\\
\cline{3-12}
& & I & T & R10 & R50 & R10 & R50 & R10 & R50 & R10 & R50 \\
\hline
\multirow{8}{*}{ZERO-SHOT} & Image-only & - & - & 5.4 & 13.9 & 9.9 & 20.8 & 8.3 & 17.7 & 7.9 & 17.5 \\
& Text-only & - & - & 13.6 & 29.7 & 18.9 & 31.8 & 19.3 & 37.0 & 17.3 & 32.9 \\
& Image+Text & - & - & 16.3 & 33.6 & 21.0 & 34.5 & 22.2 & 39.0 & 19.8 & 35.7 \\
& Slerp \cite{jang2024spherical} & - & - & 20.3 & 40.8 & 27.2 & 42.5 & 28.1 & 46.7 & 25.2 & 43.3 \\
& Pic2Word \cite{saito2023pic2word} & 100$\%$ & \color{red}{0$\%$} & 20.0 & 40.2 & 26.2 & 43.6 & 27.9 & 47.4 & 24.7 & 43.7\\
& SEARLE \cite{baldrati2023zero} & 100$\%$ & \color{red}{0$\%$} & 20.3 & 43.2 & 27.4 & 45.7 & 29.3 & 50.2 & 25.7 & 46.3\\
& LinCIR \cite{gu2024language} & \color{red}{0$\%$} & 100$\%$ & 20.9 & 42.4 & 29.1 & 46.8 & 28.8 & 50.2 & 26.3 & 46.5\\
& KEDs \cite{suo2024knowledge} & 100$\%$ & 100$\%$ & 21.7 & 43.8 & 28.9 & 48.0 & 29.9 & 51.9 & 26.8 & 47.9 \\
\cline{2-12}
& DeG & \color{red}{10$\%$} & \color{red}{10$\%$} & \textbf{24.4} & \textbf{46.5} & \textbf{30.7} & \textbf{50.3} & \textbf{31.6} & \textbf{52.0} & \textbf{28.9} & \textbf{49.6} \\
\hline
CIRR & Combiner \cite{baldrati2022effective} & - & - & 17.2 & 37.9 & 23.7 & 39.4 & 24.1 & 43.9 & 21.7 & 40.4 \\
Fashion-IQ & Combiner \cite{baldrati2022effective} & - & - & 30.3 & 54.5 & 37.2 & 55.8 & 39.2 & 61.3 & 35.6 & 57.2 \\
Fashion-IQ & Combiner* \cite{baldrati2022effective} & - & - & 31.6 & 56.7 & 36.4 & 58.0 & 38.2 & 62.4 & 35.4 & 59.0  \\
Fashion-IQ & CIRPLANT \cite{liu2021image} & - & - & 17.5 & 40.4 & 17.5 & 38.8 & 21.6 & 45.4 & 18.9 & 41.5\\
Fashion-IQ & ARTEMIS \cite{delmas2022artemis} & - & - & 27.2 & 52.4 & 21.8 & 43.6 & 29.2 & 54.8 & 26.1 & 50.3\\
Fashion-IQ & MAAF \cite{dodds2020modality} & - & - & 23.8 & 48.6 & 21.3 & 44.2 & 27.9 & 53.6 & 24.3 & 48.8 \\
\hline
\end{tabular}
\vspace{-\baselineskip}
\end{center}
\end{table*}
\section{Experiments}
\subsection{Experimental Settings}
\quad \textbf{Datasets.} For the training dataset, we use the Conceptual Caption Three Million (CC3M) \cite{sharma2018conceptual} dataset which is the universal pretraining dataset for ZS-CIR task. It is worth noting that we are only able to download 2.7M, which is less than the original amount 3M, and we randomly select 10$\%$ amount of total training data as the training set.  For the inference dataset, we evaluate our method on four datasets, including Fashion-IQ \cite{wu2021fashion}, CIRR \cite{liu2021image}, CIRCO \cite{baldrati2023zero} and COCO \cite{lin2014microsoft} to validate the different compositional capability of our method. 

Standard ZS-CIR benchmarks: Fashion-IQ, CIRR and CIRCO are three datasets for standard ZS-CIR evaluation. Fashion-IQ concentrates on attribution manipulation that the text descriptions contain the modification of attributes from relative images to target images. CIRR focuses on scene manipulation that consists of multi-sourced images with hand-written text descriptions. CIRCO is a large-scale open-domain dataset with multiple target images for each relative image and text description. Compared with CIRR, CIRCO dataset is more noisy-free for practical usage.

Object Composition: COCO validation set consists of 5,000 images and we follow \cite{saito2023pic2word} to crop an object based on the instance mask annotations to construct the reference images. The goal is to convert the cropped reference images to the original images, with the text description on the object.

\textbf{Implementation Details.} We adopt the CLIP model \cite{radford2021learning} with ViT-L/14 as the frozen image and text encoder for a fair comparison. The optimizer is AdamW \cite{loshchilov2017decoupled} with a learning rate of 5e-4, weight decay of 0.1, and a linear warmup of 1,000 steps for 10$\%$ of the total training data. The batch size is set to 1024 on 8 RTX 4090 GPUs. The hyper-parameter $\lambda$ is set to 0.5, while $\alpha$ and $\beta$ are set to 1, 2. For evaluation, the hyper-parameter $\gamma$ is set to 0.6 for CIRR and 0.7 for CIRCO, 1 for Fashion-IQ and COCO. The evaluation metric remains identical with previous work \cite{saito2023pic2word,baldrati2023zero}. For CIRCO, we use mean Average Precision (mAP@K) scores, while for other datasets, we use the top-K retrieval (R@K) results. More details are provided in the Supplementary Material.

\subsection{Comparison with State-of-the-Art Methods}
\quad We compare the proposed DeG with the most recent vision-language pretraining methods for the ZS-CIR task, including Pic2Word \cite{saito2023pic2word}, SEARLE \cite{baldrati2023zero}, LinCIR \cite{gu2024language} and KEDs \cite{suo2024knowledge}. Moreover, quantitative comparisons between several baselines are conducted, including Image-only, Text-only, and other interpolation methods, such as Image+Text and Slerp \cite{jang2024spherical}. 
Furthermore, supervised CIR methods are reported \cite{baldrati2022effective,delmas2022artemis,liu2021image,vo2019composing,dodds2020modality} for comparison, which are trained on the in-distribution triplets of the inference dataset. Lastly, the amount of data used for training on zero-shot methods is compared, where ``I" and``T" denote the percentage of image and text of the total CC3M training set that are used for training. Experiments with different scales of the training data are provided in the Supplementary Material.

\begin{table}[ht]
\setlength{\abovecaptionskip}{1pt} 
\caption{Evaluation on CIRR test set. * indicates the supervised method Combiner with ResNet50x4 as the backbone.}
\label{tab:cirr}
\begin{center}
\setlength{\tabcolsep}{1.5pt}
\begin{tabular}{c|c|cc|ccc}
\hline
\multirow{2}{*}{Supervision} & \multirow{2}{*}{Methods} & \multicolumn{2}{c|}{Amount} & \multirow{2}{*}{R1} & \multirow{2}{*}{R5} & \multirow{2}{*}{R10} \\
\cline{3-4}
& & I & T & & & \\
\hline
\multirow{8}{*}{ZERO-SHOT} & Image-only & - & - & 7.4 & 23.6 & 34.0 \\
& Text-only & - & - & 20.9 & 44.8 & 56.7\\
& Image+Text & - & - & 12.4 & 36.2 & 49.1 \\
& Slerp \cite{jang2024spherical} & - & - & 24.6 & 50.8 & 63.5\\
& Pic2Word \cite{saito2023pic2word} & 100$\%$ & \color{red}{0$\%$} & 23.9 & 51.7 & 65.3 \\
& SEARLE \cite{baldrati2023zero} & 100$\%$ & \color{red}{0$\%$} & 24.2 & 52.5 & 66.3 \\
& LinCIR \cite{gu2024language} & \color{red}{0$\%$} & 100$\%$ & 25.0 & 53.3 & 66.7 \\
& KEDs \cite{suo2024knowledge} & 100$\%$ & 100$\%$ & 26.4 & 54.8 & 67.2 \\
\cline{2-7}
& DeG & \color{red}{10$\%$} & \color{red}{10$\%$} & \textbf{26.8} & \textbf{55.0} & \textbf{67.7} \\
\hline
CIRR & Combiner \cite{baldrati2022effective} & - & - & 30.3 & 60.4 & 73.2\\
Fashion-IQ & Combiner \cite{baldrati2022effective} & - & - & 20.1 & 47.7 & 61.6 \\
CIRR & Combiner* \cite{baldrati2022effective} & - & - & 33.6 & 65.4 & 77.4 \\
CIRR & TIRG \cite{vo2019composing} & - & - & 14.6 & 48.4 & 64.1 \\
CIRR & ARTEMIS \cite{delmas2022artemis} & - & - & 17.0 & 46.1 & 61.3 \\
CIRR & CIRPLANT \cite{liu2021image} & - & - & 19.6 & 52.6 & 68.4 \\
\hline
\end{tabular}
\vspace{-\baselineskip}
\end{center}
\end{table}

\textbf{Fashion-IQ.}
Fashion-IQ focuses on the fashion attribute manipulation that commonly appears in e-commerce platform for users. Table \ref{tab:fashion} shows the impressive results of our method on different domains, surpassing all previous SOTA methods on both Recall@10 and Recall@50 with a relatively large margin. Specifically, the average performance on Recall@10 exceeds KEDs for 2.1$\%$, including 2.7$\%$, 1.8$\%$ and 1.7$\%$ for Dress, Shirt and TopTee domain, indicating the strong generalization ability on fine-grained information within different narrow domains. Moreover, we utilize less data than all previous methods that saves computational costs but achieves better performance, demonstrating the strong practicality of our method.

\textbf{CIRR}
As shown in Table \ref{tab:cirr}, our method DeG surpasses all previous methods with less training data. Concretely, DeG exceeds the SOTA method KEDs for ZS-CIR by 0.4$\%$ on Recall@1, and 0.5$\%$ on Recall@10. More importantly, we observe that previous methods use complex models, such as LLMs in SEARLE, which still performs 2.6$\%$ worse than our method. This experimental results reveal that DeG successfully excavates the semantics even within a small amount of training dataset without the assistance of any external techniques. Additionally, compared with many supervised methods, our method uses only 10$\%$ out-of-distribution image-text pairs for training to outperform methods using in-distribution triplets for training.

\textbf{CIRCO.}
The large-scale nature of the CIRCO dataset poses challenge on previous methods that the improvement is relatively incremental. The SOTA method on CIRCO is LinCIR, which improves the baseline method SEARLE with merely 0.9$\%$ on mAP@5 and 0.8$\%$ on mAP@50. However, as presented in Table \ref{tab:circo}, our method further improves LinCIR with 1.1$\%$ and 1.8$\%$ on mAP@5 and mAP@50. Even the training data is limited to 10$\%$, our method possesses strong robustness regardless of the data scale of the inference dataset.

\begin{table}[htb]
\caption{Evaluation on CIRCO dataset.}
\vspace{-\baselineskip}

\label{tab:circo}
\begin{center}
\setlength{\tabcolsep}{0.5pt}
\begin{tabular}{c|cc|cccc}
\hline
\multirow{2}{*}{Methods} & \multicolumn{2}{c|}{Amount} & \multirow{2}{*}{mAP5} & \multirow{2}{*}{mAP10} & \multirow{2}{*}{mAP25} & \multirow{2}{*}{mAP50} \\
\cline{2-3}
& I & T & & & & \\
\hline
Image-only & - & - & 1.9 & 2.2 & 3.1 & 3.6 \\
Text-only & - & - & 2.7 & 3.0 & 3.7 & 3.9 \\
Image+Text & - & - & 5.0 & 6.1 & 7.3 & 8.0 \\
Slerp \cite{jang2024spherical} & - & - & 8.8 & 9.8 & 11.3 & 12.0\\
Pic2Word \cite{saito2023pic2word} & 100$\%$ & \color{red}{0$\%$}  & 8.7 & 9.5 & 10.6 & 11.3\\
SEARLE \cite{baldrati2023zero} & 100$\%$ & \color{red}{0$\%$} & 11.7 & 12.7 & 14.3 & 15.1\\
LinCIR \cite{gu2024language} & \color{red}{0$\%$} & 100$\%$ & 12.6 & 13.6 & 15.0 & 15.9\\
\hline
DeG & \color{red}{10$\%$} & \color{red}{10$\%$} & \textbf{13.7} & \textbf{14.9} & \textbf{16.8} & \textbf{17.7} \\
\hline
\end{tabular}
\vspace{-\baselineskip}
\end{center}
\end{table}

\textbf{COCO.}
Object composition is another line of CIR task that is a commonsense in real life to search for the target image with desired objects. As shown in Table \ref{tab:coco}, our method performs extremely well on this task that outperforms all previous methods with a large margin, 2.5$\%$ and 2.0$\%$ improvement on Recall@1 by Pic2Word and KEDs. Especially, KEDs merely improves on the baseline Pic2Word with 0.5$\%$, while our improvement is 5 times better. Moreover, DeG is even the first zero-shot method that exceeds the supervised training method trained on CIRR and Fashion-IQ, with 4.1$\%$ and 0.8$\%$.

\begin{table}[htb]
\caption{Results on COCO dataset for the object composition task.}
\vspace{-\baselineskip}
\label{tab:coco}
\begin{center}
\setlength{\tabcolsep}{2pt}
\begin{tabular}{c|c|cc|ccc}
\hline
\multirow{2}{*}{Supervision} & \multirow{2}{*}{Methods} & \multicolumn{2}{c|}{Amount} & \multirow{2}{*}{R1} & \multirow{2}{*}{R5} & \multirow{2}{*}{R10} \\
& & I & T & & & \\
\hline
\multirow{6}{*}{ZERO-SHOT} & Image-only & - & - & 8.6 & 15.4 & 18.9 \\
& Text-only & - & - & 6.1 & 15.7 & 23.5 \\
& Image+Text & - & - & 10.2 & 20.2 & 26.6 \\
& Slerp \cite{jang2024spherical} & - & - & 9.8 & 21.9 & 30.4 \\
& Pic2Word \cite{saito2023pic2word} & 100$\%$ & \color{red}{0$\%$} & 11.5 & 24.8 & 33.4 \\
& KEDs \cite{suo2024knowledge} & 100$\%$ & 100$\%$ & 12.0 & 26.0 & 34.9 \\
\cline{2-7}
& DeG & \color{red}{10$\%$} & \color{red}{10$\%$} & \textbf{14.0} & \textbf{28.7} & \textbf{37.7} \\
\hline
CIRR & Combiner \cite{baldrati2022effective} & - & -  & 9.9 & 22.8 & 32.2\\
Fashion-IQ & Combiner \cite{baldrati2022effective} & - & - & 13.2 & 27.1 & 35.2 \\
\hline
\end{tabular}
\vspace{-\baselineskip}
\end{center}
\end{table}

\subsection{Ablation Study}

\quad \textbf{Key Components.}
Table \ref{tab:component} shows the effectiveness of two novel designs. We first experiment on the baseline method Pic2Word with merely 10$\%$ data. Then, for the Textual Supplement (TS) module, the consistency loss function $\mathcal{L}_{mse}$ plays a significant role to align the original mapping network and the TS module, without which the performance severely degrades on all datasets. Meanwhile, the loss function $\mathcal{L}_{itcon}$ can improve overall performance, indicating the successful alignment between complementary textual information and visual information. As for the Semantic Set (S-Set), we present two experiments, including ``w/o all" that directly removes the S-Set, and ``w/o select" that sets $\lambda$ to zero for sample selection without the guidance of CLIP. Results validate that the semantics from S-Set is significant to improve overall performance, and choosing more semantic data via CLIP enhances generalization ability.

\begin{table}[htb]
\caption{Ablation studies on Fashion-IQ validation set and CIRCO dataset.}
\vspace{-\baselineskip}
\label{tab:component}
\begin{center}
\setlength{\tabcolsep}{1.5pt}
\begin{tabular}{c|c|cc|ccc}
\hline
\multirow{2}{*}{Mod.} & \multirow{2}{*}{Method} & \multicolumn{2}{c|}{Fashion-IQ} & \multicolumn{3}{c}{CIRCO} \\
\cline{3-7}
& & R10 & R50 & mAP5 & mAP10 & mAP25 \\
\hline
Baseline & Pic2Word \cite{saito2023pic2word} & 24.4 & 42.0 & 8.9 & 10.1 & 11.6 \\
\hline
\multirow{2}{*}{TS} & w/o $\mathcal{L}_{itcon}$ & 28.1 & 48.0 & 11.7 & 13.0 & 14.8 \\
& w/o $\mathcal{L}_{mse}$ & 25.3 & 43.9 & 9.4 & 10.3 & 11.7\\
\hline
\multirow{2}{*}{S-Set} & w/o select & 28.4 & 49.0 & 13.2 & 14.4 & 16.2 \\
& w/o all & 27.7 & 47.9 & 11.8 & 12.7 & 14.3 \\
\hline
All & DeG & \textbf{28.9} & \textbf{49.6} & \textbf{13.7} & \textbf{14.9} & \textbf{16.8} \\
\hline
\end{tabular}
\vspace{-\baselineskip}
\end{center}
\end{table}

\begin{figure}[htb]
  \centering
  \includegraphics[width=0.9\linewidth,height=0.23\textheight]{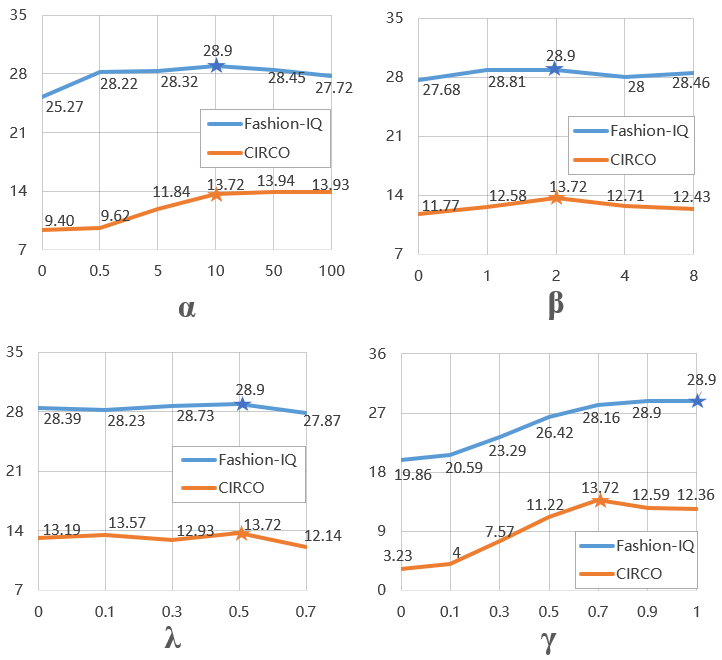}  %
    \caption{Experimental results of hyper-parameters on average performance of R@10 for Fashion-IQ validation set and mAP@5 for CIRCO dataset.}
  \label{fig:hyper}
  \vspace{-5pt}
\end{figure}

\textbf{Hyper-parameter Analysis.}
There are four hyper-parameters in our method, including $\alpha$, $\beta$ to balance loss functions and $\lambda$ to filter the semantic set during training, as well as $\gamma$ for integration during inference. As shown in Figure \ref{fig:hyper}, the trend of $\alpha$ indicates the significance of $\mathcal{L}_{mse}$ to align the mapping network with the TS module, without which the performance dramatically declines 3.6$\%$ on Fashion-IQ. The performance of $\beta$ shows that the semantics of the S-Set should neither be over-exploited that the performance drops 1.3$\%$ on CIRCO, nor entirely discarded to cause 1.9$\%$ performance drop on CIRCO. The experimental results on $\lambda$ show that the performance does not decrease monotonically with the increase of $\lambda$, indicating the necessity to choose certain semantic samples with similar captions from CLIP. For $\gamma$, the optimal value for diverse datasets is accordingly different. It can be concluded that the more complex the scene and the more compositional the caption, the smaller value of $\gamma$.

\textbf{Computational Costs.}
We evaluate the computational costs with SOTA methods from different paradigms on multiple factors, including external model usage, extra data generation, training time and inference time. As presented in Table \ref{tab:time}, compared with the pseudo-word token paradigm \cite{saito2023pic2word, gu2024language, suo2024knowledge}, our method significantly reduces training time, 33$\%$ less than the language-only training method LinCIR \cite{gu2024language}, and 20 times less than SOTA method KEDs \cite{suo2024knowledge}. Moreover, our method introduces no external models, \textit{e.g.,} Diffusion, VLMs, LLMs, which significantly reduces the inference latency. Compared with the training-free method CIReVL \cite{karthik2023vision}, our method consumes approximately 60 times less inference time per sample and also saves the cost for external models to generate captions. CompoDiff utilizes diffusion model to generate triplets in which the training time exponentially increases. In summary, our method consumes both low training overhead and inference latency that is more realistic for real-world deployment, especially deployment under asymmetric conditions between training and inference, \textit{e.g.,} the large-scale e-commerce scenario.
\begin{table}[htb]
\setlength{\abovecaptionskip}{-3pt} 
\setlength{\belowcaptionskip}{-1pt} 
\setlength{\tabcolsep}{2pt}
\caption{Comparison of computational efficiency on one A100.}
\label{tab:time}
\begin{center}
\begin{tabular}{c|c|c|c|c}
\hline
Method & External Model & Extra Data & Train & Infer \\
\hline
Pic2Word & $\times$ & $\times$ & 75h & 0.02s \\
LinCIR & $\times$ & $\times$ & 12h & 0.02s \\
KEDs & $\times$ & $\times$ & 170h & 0.03s \\
\textbf{DeG (ours)} & $\times$ & $\times$ & 8h & 0.02s \\
\hline
CompoDiff & Diffusion & 18M imgs & - & 0.23s \\
CIReVL & BLIP-2+ChatGPT & 2 caps/img & $\times$ & 1.2s \\
SEIZE & BLIP-2+GPT-4 & 30 caps/img & $\times$ & 23s \\
\hline
\end{tabular}
\vspace{-\baselineskip}
\end{center}
\end{table}
\subsection{Visualization}
\begin{figure}[ht]
  \setlength{\belowcaptionskip}{-2pt}
  \centering
  \includegraphics[width=0.9\linewidth,height=0.25\textheight]{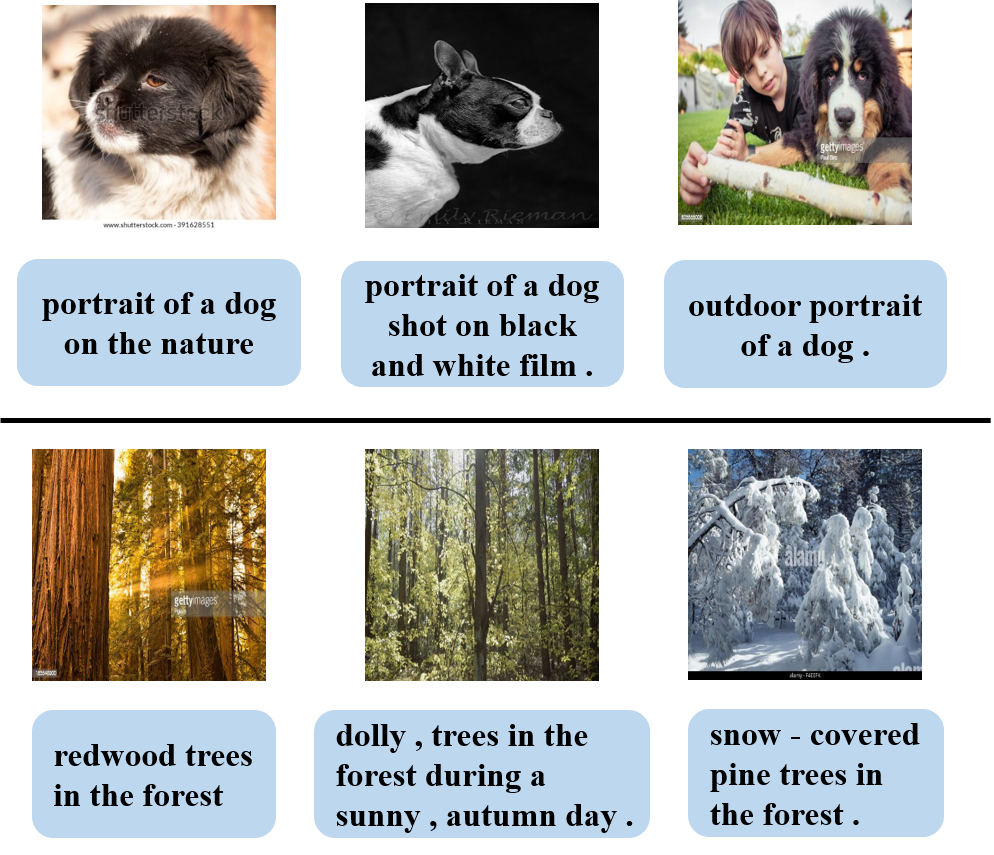}
    \caption{Visualization of the selected image-text pairs in the semantic set, which is constructed by images with similar captions to capture fine-grained information. }
  \label{fig:ss}
\vspace{-\baselineskip}
\end{figure}

\quad We visualize the images from the semantic set in Figure \ref{fig:ss}, which contains images with similar captions but have enormous differences in content. For example, ``redwood trees in the forest" and ``snow - covered pine trees in the forest" are similar from the text modality, but the images have substantial differences. Meanwhile, as the semantic set is constructed in a very flexible manner that enables self-adaptive selection in each batch, it can construct diverse negative pairs during each iteration to extract fine-grained information, which has no requirements on the original data scale or extra image database \cite{suo2024knowledge}. The visualization of the comparison between our method and others will be shown in the Supplementary Material. 

\section{Conclusion}
\quad In this paper, we observe the issues of modality discrepancy and distribution shift between training and inference. We first design a Textual Supplement module with novel training objectives to enhance the compositional understanding during training. Secondly, a Semantic Set is end-to-end mined to preserve the generalization ability from pretrained VLMs, thus improving the robustness during inference. Experimental results validate the superiority of our method on multiple datasets with much reduced training data, and further analysis shows the efficiency on computational costs.